\documentclass[letterpaper]{article} 
\usepackage{aaai25}  
\usepackage{times}  
\usepackage{helvet}  
\usepackage{courier}  
\usepackage[hyphens]{url}  
\usepackage{graphicx} 
\usepackage{natbib}  
\usepackage{caption} 
\usepackage{algorithm}
\usepackage{algorithmic}
\usepackage{amsmath}
\usepackage{amsfonts}
\usepackage{multirow}
\usepackage{multicol}
\usepackage{booktabs}
\usepackage{subfigure}
\usepackage{amssymb}
\usepackage{makecell}
\usepackage{colortbl}
\usepackage{arydshln} 
\usepackage{tcolorbox}
\usepackage{utfsym}
\usepackage{siunitx}

\usepackage{newfloat}
\usepackage{listings}
\DeclareCaptionStyle{ruled}{labelfont=normalfont,labelsep=colon,strut=off} 
\floatstyle{ruled}
\newfloat{listing}{tb}{lst}{}
\floatname{listing}{Listing}
\title{Hierarchical Divide-and-Conquer for Fine-Grained Alignment\\in LLM-Based Medical Evaluation}
\author{
    Shunfan Zheng\textsuperscript{\rm 1}, 
    Xiechi Zhang\textsuperscript{\rm 1}, 
    Gerard de Melo\textsuperscript{\rm 2}, 
    Xiaoling Wang\textsuperscript{\rm 1}, 
    Linlin Wang\textsuperscript{\rm 1}\thanks{Corresponding Author}
}
\affiliations{
    \textsuperscript{\rm 1} East China Normal University \\
    \textsuperscript{\rm 2} Hasso Plattner Institute / University of Potsdam \\
    \{sfzheng, 51255901060\}@stu.ecnu.edu.cn, \{xlwang, llwang\}@cs.ecnu.edu.cn, demelo@uni-potsdam.de
}

\usepackage{bibentry}
\nocopyright
\begin{document}

\maketitle

\begin{abstract}
In the rapidly evolving landscape of large language models (LLMs) for medical applications, ensuring the reliability and accuracy of these models in clinical settings is paramount. Existing benchmarks often focus on fixed-format tasks like multiple-choice QA, which fail to capture the complexity of real-world clinical diagnostics. Moreover, traditional evaluation metrics and LLM-based evaluators struggle with misalignment, often providing oversimplified assessments that do not adequately reflect human judgment. To address these challenges, we introduce HDCEval\footnote{Models and supplementary materials: https://huggingface.co/\\collections/AAAzsf/hdceval-6762cda19a07c157778aa22d}, a \textbf{H}ierarchical \textbf{D}ivide-and-\textbf{C}onquer \textbf{Eval}uation framework tailored for fine-grained alignment in medical evaluation. HDCEval is built on a set of fine-grained medical evaluation guidelines developed in collaboration with professional doctors, encompassing Patient Question Relevance, Medical Knowledge Correctness, and Expression. The framework decomposes complex evaluation tasks into specialized subtasks, each evaluated by expert models trained through Attribute-Driven Token Optimization (ADTO) on a meticulously curated preference dataset. This hierarchical approach ensures that each aspect of the evaluation is handled with expert precision, leading to a significant improvement in alignment with human evaluators.
\end{abstract}

%

\section{Introduction}

With the rapid development of large language models (LLMs) in the medical field, a range of advanced medical LLMs have been developed. However, the reliability and effectiveness of these models must be rigorously evaluated to ensure accurate and safe clinical decisions. 

\begin{figure}[t]
    \centering
    \includegraphics[width=0.95\linewidth]{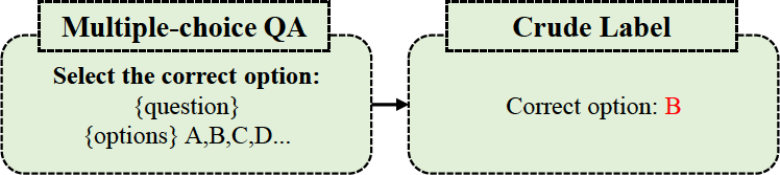}
    \caption{Fixed format task for evaluation.}
    \label{fig:data1}
    \vspace{0.5em}
    \includegraphics[width=0.95\linewidth]{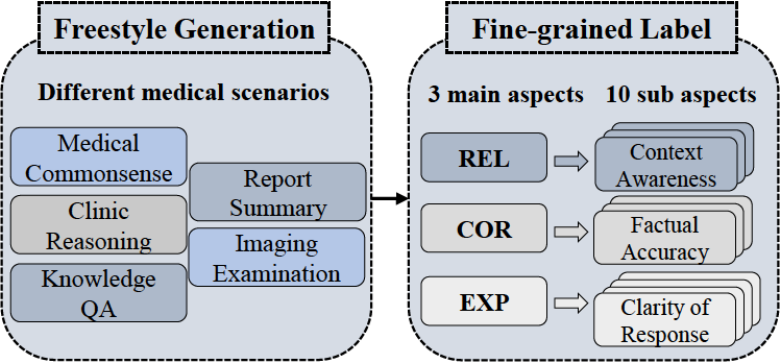}
    \caption{Freestyle fine-grained medical data for evaluation.}
    \label{fig:data2}
    \vspace{-1.5em}
\end{figure}

However, existing benchmarks such as MT-Bench \cite{zheng2024judging} and MedBench \cite{cai2024medbench} are often limited to tasks in fixed formats such as multiple-choice question answering (QA), as shown in Figure \ref{fig:data1}, lacking clinical freestyle generation, which does not align with the actual clinical diagnostic process. Moreover, current evaluation metrics fail to provide comprehensive evaluation results, instead offering only simplistic assessments. For instance, traditional n-gram metrics like ROUGE \cite{lin2004rouge} and BERT-based semantic similarity metrics \cite{zhang2019bertscore} yield only a single value, devoid of specific logical explanations.

LLMs can serve as evaluators \cite{fu2023gptscore,kocmi2023large} in such freestyle contexts due to their generative capabilities. Unlike traditional metrics, LLMs can offer more nuanced and context-aware assessments by generating detailed feedback and explanations. This allows them to better reflect complex scenarios, such as those found in clinical diagnostics. However, existing LLM evaluators often exhibit misalignment with human evaluators in medical evaluation. For instance, evaluation using GPT-4 or the open-source model PandaLM \cite{wang2023pandalm} can inadvertently perpetuate or even amplify existing bias in the training data, leading to skewed and inconsistent assessments \cite{stureborg2024large,wang2023large} that may not accurately reflect diverse patient populations or medical scenarios compared to human physicians. 

To address the issues above, we first collaborate with professional doctors to propose a set of fine-grained medical evaluation guidelines tailored for detailed medical assessments. These guidelines include three primary aspects: Patient Question Relevance (REL), Medical Knowledge Correctness (COR), and Expression (EXP), each further subdivided into specific sub-aspects.

Based on the guidelines, we introduce HDCEval, a hierarchical divide-and-conquer evaluation framework that consists of two main components. Firstly, the \textbf{Divide} component involves a hierarchical decomposition of the evaluation task. The process begins by dividing the complex evaluation into multiple primary tasks. Each primary task is then further subdivided into more detailed subtasks. For each primary task and its corresponding subtasks, we employ a specialized expert model to carry out the evaluation, ensuring precise and expert-aligned assessments. This is in contrast to the BSM method \cite{saha2023branch}, which relies on a single, non-specialized model to handle all primary tasks. 

In the \textbf{Conquer} component, the framework leverages a carefully constructed preference dataset, which is specifically designed to improve alignment with human evaluators. This dataset plays a crucial role in enhancing the performance of each expert model. Based on this dataset, we introduce the Attribute-Driven Token Optimization (ADTO) method for training. This method incorporates reward tokens that guide the optimization of different expert models, ensuring that each model aligns with the specific evaluation criteria of its assigned tasks, thereby enhancing the precision and quality of the overall evaluation. The experimental results demonstrate that HDCEval significantly outperforms existing baseline methods across various medical scenarios. Notably, compared to the PandaLM evaluator, HDCEval achieves an overall improvement in consistency with human evaluations by 23.92\%. This highlights the effectiveness of the Hierarchical Divide-and-Conquer Evaluation Framework in aligning model evaluations with expert-level assessments in the medical domain.

Our key contributions can be summarized as follows:
\begin{itemize}
    \item We propose a comprehensive set of fine-grained medical evaluation guidelines developed in collaboration with professional doctors.
    \item We introduce HDCEval, a hierarchical divide-and-conquer evaluation framework designed for detailed and accurate medical evaluations, achieving finer-grained evaluation that better aligns with human evaluators.
    \item We develop and apply the Attribute-Driven Token Optimization (ADTO) strategy, demonstrating that HDCEval surpasses other baselines in accuracy and alignment with human evaluators in freestyle medical contexts.
\end{itemize}

\section{Methodology}
\label{methodology}

\subsection{Task Formulation}
\label{sec:task_formulation}
In evaluation tasks, the input \(x\) consists of a question \(q\) and the model's response \(r\). The goal is to generate an evaluation result \(E\) composed of multiple dimensions. In our medical evaluation tasks, the final evaluation consists of \(m\) distinct dimensions, denoted as \(E= \{E_1,\ldots, E_m\}\), where
each \(E_i\) represents the assessment of a specific dimension. Each \(E_i\) is defined as a tuple
\begin{equation}
    E_i = (s_i,p_i),
\end{equation}
where \(s_i\) is the scoring of the response on dimension \(i\), and \(p_i\) is the corresponding rationale explaining the reasoning process.

\subsection{Fine-grained Medical Evaluation Guidelines}
\label{sec:fine_grained_eval}
Achieving accurate and nuanced evaluations is crucial in clinical diagnostics to ensure patient safety and effective treatment. To address this, we collaborated with medical experts to develop detailed evaluation guidelines specifically designed for medical assessments. These guidelines emphasize three primary aspects:
\begin{itemize}
    \item \textbf{Patient Question Relevance (REL):} This aspect considers how well the medical response addresses the patient's specific questions and concerns. It involves assessing the clarity, directness, and appropriateness of the response in relation to the patient's query.

    \item \textbf{Medical Knowledge Correctness (COR):} This aspect ensures the accuracy of the medical information provided. It involves evaluating whether the response aligns with current medical knowledge, guidelines, and evidence-based practices.

    \item \textbf{Expression (EXP):} This aspect focuses on the clarity and coherence of the response, assessing the language, structure, and presentation of the information to ensure it is easily understandable and professional.

\end{itemize}
Each primary aspect is further divided into 3-4 sub-aspects to capture the intricacies of medical evaluations thoroughly. For instance, Patient Question Relevance (REL) includes sub-aspects such as Relevance to Patient's Condition (COND), which assesses how directly the response pertains to the patient's specific medical condition. Medical Knowledge Correctness (COR) encompasses sub-aspects like Factual Accuracy (ACC), ensuring the information aligns with current evidence-based practices. These sub-aspects provide a granular framework for evaluation, ensuring comprehensive coverage of each aspect. For each sub-aspect, scores range from 0 to 5, with detailed scoring rules provided in the Technical Appendix within supplementary materials.

\subsection{Hierarchical Divide-and-Conquer Evaluation Framework}
\label{sec:framework}
\subsubsection{Overview}
\begin{figure*}[t]
    \centering
    \includegraphics[width=\linewidth]{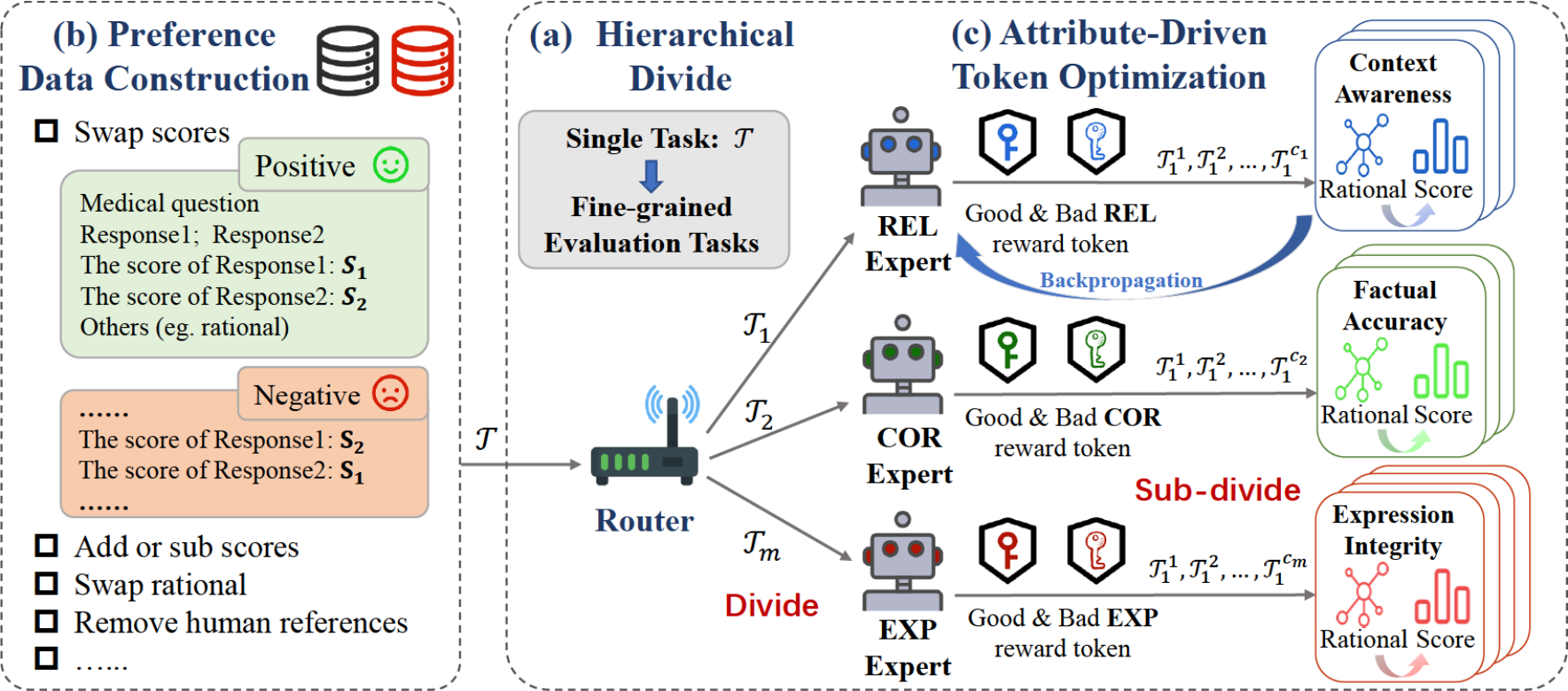}
    \caption{Overview of the Hierarchical Divide-and-Conquer Evaluation Framework. ``Hierarchical Divide'' represents the \emph{Divide} component, while ``Preference Data Construction'' and ``Attribute-Driven Token Optimization'' constitute the \emph{Conquer} component.}
    \label{fig:framework}
    \vspace{-1.0em}
\end{figure*}

As shown in Figure \ref{fig:framework}, the Hierarchical Divide-and-Conquer Evaluation Framework tackles medical evaluations by first \emph{dividing} the task into detailed, expert-focused subtasks. Then, it \emph{conquers} these tasks using preference data and Attribute-Driven Token Optimization (ADTO) to refine the model. This method ensures thorough and precise alignment with medical evaluation standards.

\subsubsection{Hierarchical Divide}
Our medical evaluation guidelines are inherently multi-dimensional and strictly constrained, making accurate assessment a challenging task. LLMs often struggle with completing nuanced guideline-based estimation tasks due to their generalized training and lack of fine-tuned specialization.

To address these challenges, we propose a hierarchical divide-and-conquer approach, inspired by BSM \cite{saha2023branch}.  BSM's methodology demonstrates the efficacy of decomposing complex evaluation tasks into manageable subtasks that can be addressed in parallel.  However, BSM's approach relies on a single model for all subtasks, which limits its ability to achieve fine-grained alignment with human evaluators.  

In contrast, our framework enhances this approach by using specialized expert models for different aspects of the evaluation. We first decompose the overarching evaluation task \(\mathcal{T}\) into \(n\) primary evaluation tasks \(\mathcal{T}_1,\ldots,\mathcal{T}_n\), each aligned with an expert model. These primary tasks are further subdivided into subtasks to capture the intricacies of the evaluation criteria.

The hierarchical decomposition is structured as follows:
\begin{equation}
\left\{
\begin{aligned}
    & \mathcal{T} = \mathcal{T}_1(x,I_1), \ldots, \mathcal{T}_m(x,I_m) \\
    & \mathcal{T}_i(x,I_i) = \mathcal{T}_i^1(x,I_i,I_i^1), \ldots, \mathcal{T}_i^c(x,I_i,I_i^{c_i})
\end{aligned}
\right.
\end{equation}
Here, \(I_i\) represents the instruction of primary evaluation tasks \(\mathcal{T}_i\) and \(I_i^j\) represents the instruction of subtasks \(\mathcal{T}_i^j\). 

Each expert model is trained specifically for one primary aspect (including its associated sub-aspects) to ensure fine-grained and accurate alignment with medical evaluation guidelines. This specialization allows for more precise evaluations that closely align with human expertise. Our hierarchical approach effectively manages the complexity of medical evaluations, ensuring a detailed and accurate assessment of each aspect.

\subsubsection{Preference Data Construction}
To enhance model alignment with human evaluators in medical assessments, we develop a preference dataset that specifically targets misalignment and bias. This dataset construction is intricately linked to our fine-grained evaluation guidelines, ensuring that model improvements align with detailed evaluation criteria. The negative samples are constructed from existing positive samples, and this process is represented by the following formula:

\begin{itemize}
    \item \textbf{Swapping Scores of Two Responses:} Let \(R_1\) and \(R_2\) be two responses with scores \(S(R_1)\) and \(S(R_2)\) from a positive sample. The scores of the corresponding negative sample are swapped: 
    \begin{equation}
        S'(R_1)=S(R_2), S'(R_2)=S(R_1)
    \end{equation}
    This method forces the model to determine which response better addresses the patient's query, refining its ability to assess relevance.
    \item \textbf{Simultaneously Adding or Subtracting Scores from Two Responses:} Considering two responses \(R_1\) and \(R_2\) with scores \(S(R_1)\) and \(S(R_2)\), we adjust the scores by a constant \(\Delta S\):
    \begin{equation}
        S'(R_1) = S(R_1) + \Delta S, \quad S'(R_2) = S(R_2) - \Delta S
    \end{equation}
    This technique helps the model differentiate between high-quality and low-quality responses by teaching it to discern changes in accuracy and presentation.
    \item \textbf{Exchanging Rationales of Two Responses:} Let \(R_1\) and \(R_2\) be two responses with corresponding rationales \(P(R_1)\) and \(P(R_2)\). We swap their rationales:
    \begin{equation}
        P'(R_1) = P(R_2), \quad P'(R_2) = P(R_1)    
    \end{equation}
    
    This method ensures that the model’s explanations align with its judgments, thereby reducing logical inconsistencies.
    \item \textbf{Removing Human-Provided Reference Information:} Let \(E_h\) represent an evaluation result that includes human-provided reference information \(I_h\). The human-provided information is removed from the evaluation result:
    \begin{equation}
        E'_h = E_h \setminus I_h
    \end{equation}
    This strategy reinforces the importance of human-provided information, allowing the model's outputs to better align with human expectations.
\end{itemize}

\subsubsection{Attribute-Driven Token Optimization}
\label{sec:ADTO}
To further reduce the bias of the evaluator, and improve the alignment between the models and professional physicians, we introduce the Attribute-Driven Token Optimization (ADTO) method.

\begin{algorithm}[t]
\renewcommand{\algorithmicrequire}{\textbf{Input:}}
\renewcommand{\algorithmicensure}{\textbf{Output:}}
\caption{Attribute-Driven Token Optimization (ADTO)}
\label{alg}
\begin{algorithmic}[1]
\REQUIRE $\mathcal{D} = \{(x_j, y_{j,w}, y_{j,l})\}_{j=1}^N$, multi-dimensional evaluation data with positive ($y_w$) and negative ($y_l$) examples, $\mathcal{M} = \{M_1, M_2, \ldots, M_m\}$, set of fine-grained evaluation models, $\mathcal{R} = \{R_{\text{REL}}, R_{\text{COR}}, R_{\text{EXP}}\}$, set of reward tokens for relevance, correctness, and expression.
\ENSURE Fine-Grained Medical Evaluation Models $\mathcal{M}$.

\STATE \textbf{Initialization:} Set model parameters $\theta_i$ for each $M_i \in \mathcal{M}$.
\FOR{each $M_i$ in $\mathcal{M}$}
    \FOR{each training step $t$}
        \STATE Sample a mini-batch $(x_j, y_{j,w}, y_{j,l})$ from $\mathcal{D}$
        \STATE Determine aspect $a \in \{\text{REL}, \text{COR}, \text{EXP}\}$ for $M_i$
        \STATE Construct reward token $r = R_{a}$
        \STATE Create inputs $z_w = \text{combine}(x_j, r, y_{j,w})$ and $z_l = \text{combine}(x_j, r, y_{j,l})$
        \STATE Compute model outputs $o_{j,w} = M_i(z_w; \theta_i)$ and $o_{i,l} = M_i(z_l; \theta_i)$
        \STATE Compute loss $\mathcal{L}(o_{j,w}, o_{j,l})$ using Attribute-Driven Token Optimization
        \STATE Freeze unrelated layers to stabilize training
        \STATE Update other parameters $\theta_i$ via gradient descent: $\theta_i \leftarrow \theta_i - \eta \nabla_{\theta_i} \mathcal{L}(o_{j,w}, o_{j,l})$
    \ENDFOR
\ENDFOR
\RETURN $\mathcal{M}$
\end{algorithmic}
\end{algorithm}

The ADTO method leverages preference datasets by embedding specific reward tokens within the training data. These tokens represent different aspects of evaluation quality, and guide the model in distinguishing between superior and inferior responses. The integration of reward tokens enables the model to learn from nuanced distinctions that are critical in professional medical assessments.

For each \(i\)-th primary aspect evaluation model in our framework, the optimization process is designed to balance the current policy model's responses with those of a reference model. This process is mathematically formulated in Eq.~\ref{eq:ADTO_i} with respect to training model parameters \(\pi_{\theta}^i\), reference model parameters \(\pi_{ref}^i\), and hyperparameter \(\beta_i\) as
\begin{equation}
\begin{aligned}
\mathcal{L}_{\mathrm{ADTO}}^i(\pi_{\theta}^{i};\pi_{\mathrm{ref}}^{i}) &= -\mathbb{E}_{(x,y_{w}^{i},y_{l}^{i})\thicksim\mathcal{D}} \Bigg[ \\
&\quad \log\sigma\left(\beta_{i} \log \frac{\pi_{\theta}^{i}(y_{w}^{i}\mid x,t_{w}^{i},I_{i})}{\pi_{\mathrm{ref}}^{i}(y_{w}^{i}\mid x,t_{w}^{i},I_{i})} \right. \\
&\quad \left. - \beta_{i} \log \frac{\pi_{\theta}^{i}(y_{l}^{i}\mid x,t_{l}^{i},I_{i})}{\pi_{\mathrm{ref}}^{i}(y_{l}^{i}\mid x,t_{l}^{i},I_{i})} \right) \Bigg],
\end{aligned}
\label{eq:ADTO_i}
\end{equation}
where \((x,y_w^i,y_l^i)\) refers to the triplet of (input, good evaluation, bad evaluation), and \(t_w^i, t_l^i\) represent different reward tokens in optimization. Here, \(\pi_{\theta}^i(y_{w}^{i}\mid x,t_{w}^{i},I_{i})\) denotes the cumulative probability of the current policy model generating good responses, while \(\pi_{\mathrm{ref}}^i(y_{l}^{i}\mid x,t_{l}^{i},I_{i})\) represents the cumulative probability of the reference model generating bad responses. \(\sigma\) denotes the sigmoid function. Then, we integrate the optimization processes of all \(m\) models within the framework. Further details are specified in Algorithm \ref{alg}.

The factual knowledge within large language models is often injected into deeper layers. Therefore, to equip the model with more accurate and objective evaluation capabilities while reducing the computational cost, we freeze the first 24 layers of our model and only train the last eight layers.

\section{Experiments}

\subsection{Experimental Setup}
\label{sec:exp_set}
Our evaluation models are based on the MedLlama2-7B model. We train using a batch size of 128 and a maximum token length of 4,096 on 4 NVIDIA A100-80GB GPUs. To maximize GPU memory usage and accelerate training, we employed the Fully Sharded Data Parallel \cite{zhao2023pytorchfsdp} strategy and the FlashAttention \cite{dao2022flashattention} algorithm. The learning rates for the instruction tuning and direct preference optimization phases are set to \num{2e-5} and \num{5e-7}, respectively. During inference, we use greedy decoding with a temperature of 0 to minimize randomness.

\subsection{Medical Dataset}
\subsubsection{Data Source}
First, we integrate medical questions from different sources including medical meadow wikidoc\footnote{https://huggingface.co/datasets/medalpaca/medical\_meadow\_\\wikidoc}, MedBench \cite{cai2024medbench}, MedText\footnote{https://huggingface.co/datasets/BI55/MedText}, and MedDialog \cite{zeng2020meddialog}. We perform automated and manual filtering to ensure reliable and safe medical question sources.
Then, to diversify the task types of the data and conform to the clinical medical scenario, we divide the data into five specific medical scenarios shown in Figure \ref{fig:data2}.

\begin{table*}[t]
\centering
\setlength{\tabcolsep}{10pt}
\begin{tabular}{lccc:ccc:cc}
    \toprule
         \multirow{2}[0]{*}{\textbf{Medical Scenarios}} & \multicolumn{3}{c}{\textbf{Pairwise Accuracy (\%)}} & \multicolumn{3}{c}{\textbf{Reference Match (\%)}} & \multicolumn{2}{c}{\textbf{Correlation}} \\
         \cmidrule(lr){2-4}\cmidrule(lr){5-7}\cmidrule(lr){8-9}
          & REL   & COR   & EXP      & REL   & COR   & EXP   & Pearson & ICC \\

    \rowcolor{gray!26}
    \multicolumn{2}{l}{\textbf{Imaging Examination (Text)}}        &       &       &       &       &       &       &  \\
    MedLlama2  & 61.54$^{*}$ & 52.38$^{*}$ & 60.91$^{*}$ & 60.72$^{*}$ & 50.24$^{*}$ & 60.86$^{*}$ & 0.5484$^{*}$ & 0.5893$^{*}$ \\
    PandaLM  & 54.15$^{*}$ & 50.53$^{*}$ & 50.53$^{*}$ & 55.84$^{*}$ & 48.49$^{*}$ & 53.71$^{*}$ & 0.5604$^{*}$ & 0.6141$^{*}$\\
    ChatGPT    & 69.23$^\dagger$  & 64.10$^\dagger$ & 55.77$^{*}$ & 70.12$^\dagger$ & 64.40$^\dagger$ & 75.97$^{*}$ & 0.5813$^\dagger$ & 0.6693$^\dagger$ \\
    GPT-4         & \textbf{84.87$^{*}$}  & 61.54$^\dagger$  & 71.15$^{*}$ & \textbf{80.74$^{*}$} & 69.46$^\dagger$ & 88.41$^{*}$ & 0.5898$^\dagger$ & 0.6849$^\dagger$ \\
    Ours (HDCEval) & \textbf{84.87$^{*}$}  & \textbf{79.49$^{*}$}  & \textbf{75.00$^{*}$}  &  78.78$^{*}$  & 70.03$^{*}$  & \textbf{92.98$^{*}$}  &  \textbf{0.6480$^{*}$} & \textbf{0.7149$^{*}$} \\

    \rowcolor{gray!26}
    \textbf{Clinic Reasoning} &       &       &       &       &       &       &       &  \\
    MedLlama2  & 57.14$^{*}$ & 48.83$^{*}$ & 56.67$^{*}$ & 59.72$^{*}$ & 51.43$^{*}$ & 56.32$^{*}$ &  0.4060$^{*}$ & 0.5247$^{*}$
    \\
    PandaLM  &  50.61$^{*}$ &  46.12$^{*}$ & 47.29$^{*}$ & 56.67$^{*}$ & 39.25$^{*}$ & 51.62$^{*}$ & 0.3637$^{*}$ & 0.4919$^{*}$ \\
    ChatGPT    & 64.63$^\dagger$ & 64.63$\S$ & 59.69$^\dagger$ & 66.56$^\dagger$ & 64.74$\S$ & 67.25$^\dagger$ & 0.5483$\S$ & 0.7101$\S$ \\
    GPT-4            & 78.91$^{*}$  &	69.18$^\dagger$  & 60.20$^{*}$ & 78.91$^\dagger$  & 70.44$^\dagger$ & 75.13$^\dagger$ & 0.5599$^\dagger$ & \textbf{0.7209$^\dagger$} \\
    Ours (HDCEval) & \textbf{82.99$^{*}$}  & \textbf{67.35$^{*}$}  & \textbf{67.35$^{*}$}  &  \textbf{88.23$^{*}$}  & \textbf{82.50$^{*}$}  & \textbf{85.81$^{*}$}  & \textbf{0.5887$^{*}$} &  \textbf{0.7209$^{*}$} \\

    \rowcolor{gray!26}
    \textbf{Knowledge QA} &       &       &       &       &       &       &       &  \\
    MedLlama2  & 56.67$^{*}$ & 47.08$^{*}$ & 52.61$^{*}$ & 56.67$^{*}$ & 50.73$^{*}$ & 54.52$^{*}$ & 0.5240$^{*}$ & 0.6917$^{*}$ \\
    PandaLM  & 40.53$^{*}$ & 39.18$^{*}$ & 41.55$^{*}$ & 48.65$^{*}$ & 40.11$^{*}$ & 45.74$^{*}$ & 0.5181$^{*}$ & 0.6469$^{*}$ \\
    ChatGPT    & 63.33$^{*}$ & 71.11$^{*}$ & 58.33$^{*}$ & 68.35$^{*}$ & 72.01$^{*}$ & 70.57$^\dagger$ & 0.5603$^{*}$ & 0.6519$^{*}$ \\
    GPT-4      & 76.11$^{*}$ & 66.11$^{*}$ & 61.67$^{*}$ & 79.86$^{*}$ & 73.67$^{*}$ & 73.86$^{*}$ & 0.5632$^{*}$ & 0.6656$^{*}$ \\
    Ours (HDCEval) &  \textbf{85.00$^{*}$}  & \textbf{73.33$^{*}$}  & \textbf{74.17$^{*}$}  & \textbf{86.78$^{*}$}  & \textbf{78.41$^{*}$}  & \textbf{81.85$^{*}$}  & \textbf{0.5693$^{*}$} &  \textbf{0.7073$^{*}$}   \\

    \rowcolor{gray!26}
    \textbf{Report Summary} &       &       &       &       &       &       &       &  \\
    MedLlama2  & 60.85$^{*}$ & 58.86$^{*}$ & 61.28$^{*}$ & 61.63$^{*}$ & 58.36$^{*}$ & 62.96$^{*}$ & 0.4303$^{*}$ & 0.5595$^{*}$ \\
    PandaLM    & 58.47$^{*}$ & 45.20$^{*}$ & 62.07$^{*}$ & 62.79$^{*}$ & 50.19$^{*}$ & 63.46$^{*}$ & 0.3947$^{*}$ & 0.5342$^{*}$ \\
    ChatGPT    & 72.13$^\dagger$ & 66.24$^{*}$ & 69.68$^{*}$ & 77.01$^{*}$ & 67.66$^\dagger$ & 73.72$^{*}$ & 0.5864$^\dagger$ & 0.6936$^\dagger$ \\
    GPT-4      & 74.88$^{*}$ & 70.10$^{*}$ & \textbf{70.41$^{*}$} & 77.06$^\dagger$ & 68.84$^{*}$ & \textbf{75.71$^{*}$} & 0.5905$^{*}$ & 0.6948$^{*}$ \\
    Ours (HDCEval) & \textbf{75.24$^{*}$} & \textbf{72.76$^{*}$} & 70.03$^{*}$ & \textbf{77.62$^{*}$} & \textbf{72.31$^{*}$} & 73.75$^{*}$ & \textbf{0.5913$^{*}$} & \textbf{0.7047$^{*}$} \\
    
    \rowcolor{gray!26}
    \multicolumn{2}{l}{\textbf{Medical Commonsense}}   &       &       &       &       &       &       &  \\
    MedLlama2  & 58.82$^{*}$ & 49.41$^{*}$ & 56.73$^{*}$ & 61.88$^{*}$ & 53.92$^{*}$ & 56.13$^{*}$ & 0.3609$^{*}$ & 0.4923$^{*}$ \\
    PandaLM  & 57.05$^{*}$ & 41.53$^{*}$ & 52.71$^{*}$ & 60.57$^{*}$ & 43.63$^{*}$ & 54.12$^{*}$  & 0.3256$^{*}$ & 0.4507$^{*}$ \\
    ChatGPT    & 70.59$^{*}$ & 68.55$^{*}$ & 61.77$^{*}$ & 71.41$^{*}$ & 72.94$^{*}$ & 68.33$^\dagger$ & 0.5815$^{*}$ & 0.7238$^{*}$ \\
    GPT-4      & 72.55$^{*}$ & 68.63$^{*}$ & \textbf{79.41$^{*}$} & 74.25$^\dagger$ & 76.91$^{*}$ &  69.88$^{*}$ & 0.5954$^{*}$ & 0.7236$^{*}$ \\
    Ours (HDCEval) & \textbf{74.51$^{*}$}  & \textbf{70.59$^{*}$}  & 77.94$^{*}$  &  \textbf{76.58$^{*}$}  & \textbf{88.35$^{*}$}  & \textbf{71.50$^{*}$}  & \textbf{0.6767$^{*}$} &  \textbf{0.7881$^{*}$}\\

    
    \bottomrule
\end{tabular}
\caption{Fine-grained evaluation results. We run models three times and report the average results. * represents a significant difference with our results or significant correlation with human evaluation (t-test, \emph{p}-value$<$ 0.001), while $\dagger$ and $\S$
refer to t-test with  \emph{p}$<$0.01 and \emph{p}$<$0.05, respectively.}
\label{tab:Main Results}
\vspace{-1.3em}
\end{table*}

\subsubsection{Dataset Construction}
First, we use four different medical models: ChatDoctor, Baize, MedAlpaca, and MedLlama2, to generate responses to the questions. Then, with the assistance of AI, we annotated the 13,452 samples following the Guidelines Instructions. More details about the dataset construction are provided in the Technical Appendix within supplementary materials.

\subsubsection{Dataset Validation}
To validate the effectiveness of our dataset, we use the publicly available MedMCQA dataset \cite{pal2022medmcqa} as a reference. We evaluated four models on both datasets and calculated their rankings\footnote{https://github.com/ctlllll/understanding\_llm\_benchmarks}. The results show consistent rankings of the models across the two datasets, with ChatDoctor demonstrating the best performance. Additionally, we find that ChatDoctor exhibits the strongest ability to follow instructions.

\subsection{Baselines and Test Set}
We selected representative models as baselines, including the closed-source models (ChatGPT and GPT-4) and the open-source models (PandaLM and MedLlama2). 

For the test data, we initially extracted 2,994 samples from the constructed dataset to form the test set, with the remaining samples used as the training set. We then hired human doctors to annotate the test data. The annotation process follows the fine-grained medical evaluation guidelines.

\subsection{Evaluation}
The generated evaluation results include both scores and rationales. Therefore, we need to assess these two aspects separately. For the scores, we employ automated metrics, while for the rationales, we rely on evaluations conducted by human doctors.

\paragraph{Human Evaluation} The human annotators manually assess whether the rationale from the model matches the rationale from human-provided labels to verify the reasonableness of the model's evaluation results. This process is referred to as \textbf{Reference Match}. If the label's rationale indicates an error in the medical knowledge within the response, but the model fails to recognize it, it is considered a mismatch.

\paragraph{Automatic Metrics} We use \textbf{Pairwise Accuracy} as the primary evaluation metric for the scores. If the relative ranking of the evaluation scores between the two responses generated by the model is consistent with the labels from human doctors, it indicates that the model accurately evaluated the quality of the two responses; otherwise, it does not.
Additionally, we use the \textbf{Intraclass Correlation Coefficient} \cite{koo2016guidelineicc} and \textbf{Pearson Correlation Coefficient} \cite{cohen2009pearson} to measure the similarity between the model evaluation and human evaluation.

\subsection{Main Results}

\begin{table*}[t!]
    \centering
    
    \begin{tabular}{lccccccccccc}
    \toprule
           \multirow{2}[0]{*}{\textbf{Method}} & \multicolumn{3}{c}{\textbf{REL}} & \multicolumn{3}{c}{\textbf{COR}} & \multicolumn{4}{c}{\textbf{EXP}} & \multirow{2}[0]{*}{\textbf{AVG}} \\
           \cmidrule(lr){2-4}\cmidrule(lr){5-7}\cmidrule(lr){8-11}
                 & CONT    & COND  & CONC   & ACC    & INFO  & UNC    & CLAR   & LANG    & TE   & INTE    &  \\
    \midrule
        HDCEval& \textbf{80.91} & \textbf{81.82} & \textbf{78.83} & \textbf{71.65} & \textbf{72.18 } & \textbf{74.27}  & \textbf{73.48 } & \textbf{75.14} & \textbf{70.84 } & \textbf{72.14 } & \textbf{75.13 } \\
           
        HDCEval$_{\text{no-token}}$  & 78.57  & 79.50  & 78.36  & 69.93  & 68.12  & 68.51  & 68.40  & 70.81  & 67.89  & 69.98  & 72.01  \\

        HDCEval$_{\text{no-preference}}$ & 80.70  & 80.97  & 77.94  & 71.35  & 69.57  & 71.29 & 71.36  & 69.80  & 67.13  & 71.17  & 73.13  \\

    \bottomrule
    \end{tabular}
    \caption{Ablation Study on HDCEval Components -- Assessing the impact of removing reward tokens and preference data on evaluation accuracy.}
    \label{tab:ablation}
    \vspace{-1.3em}
\end{table*}

\paragraph{Evaluation Metrics Results} To demonstrate the capability of HDCEval in fine-grained medical assessment, we arranged for medical experts to evaluate the responses. Table \ref{tab:Main Results} provides the fine-grained evaluation results of HDCEval compared to the baselines. From left to right, the results of three different metrics on fine-grained dimensions are included. We observe that HDCEval outperforms other models across multiple scenarios, especially outperforming GPT-4 on reference match and correlation metrics, which reflects better alignment with humans. Regarding pairwise accuracy metrics, it demonstrates a 23.92\% improvement compared to PandaLM. From the fine-grained perspective, HDCEval significantly improves evaluation accuracy compared to other baselines in terms of Medical Knowledge Correctness (COR).

\begin{figure}[h!]
    \centering
    \includegraphics[width=\linewidth]{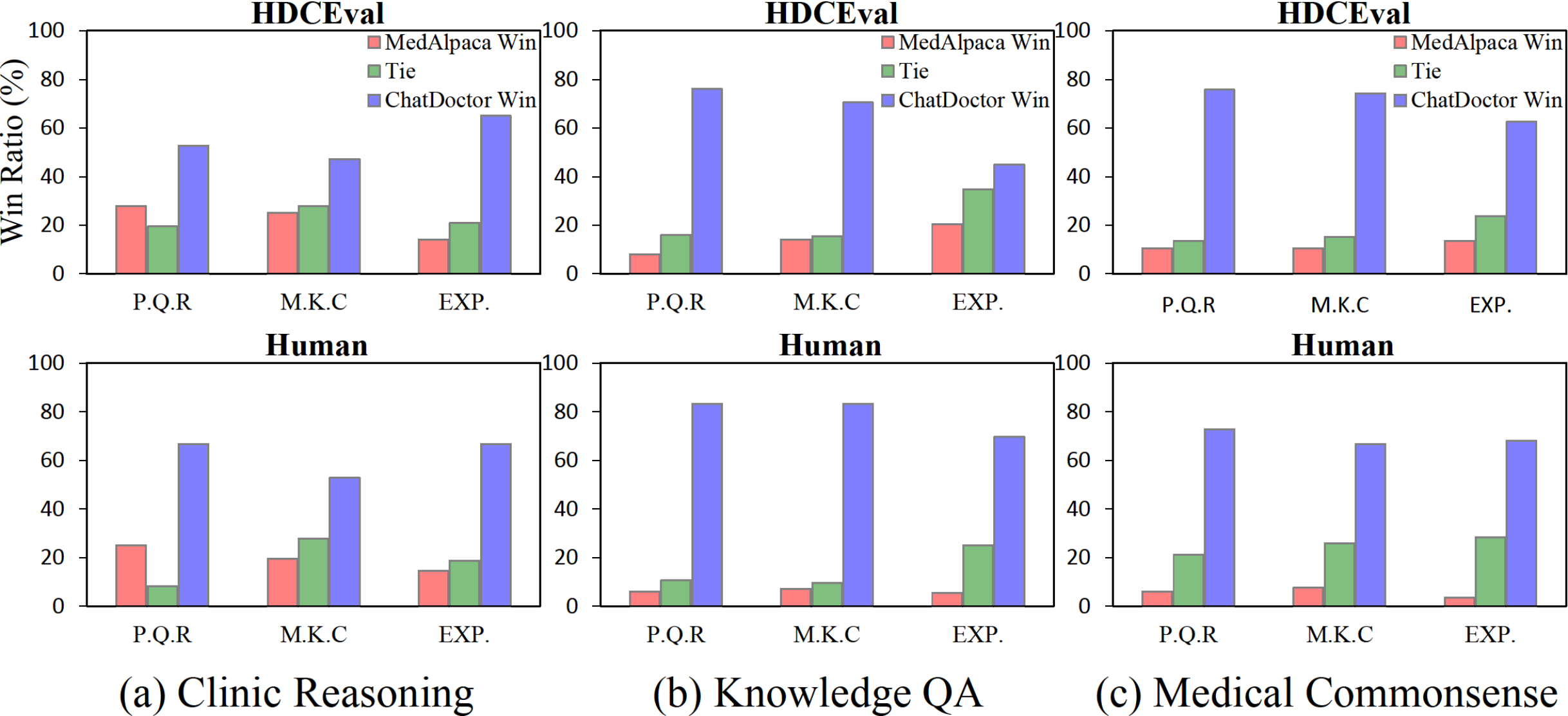}
    \caption{The performance of MedAlpaca and ChatDoctor across multiple medical scenarios is evaluated using HDCEval and compared to human doctors' judgments. ``Win'' indicates the percentage of cases where a given medical language model outperforms the other, while ``Tie'' indicates the percentage of cases where both medical LLMs received the same score.}
    \label{fig:win_tie_lose_domain}
\end{figure}

\paragraph{Win-Tie-Lose Experiment Results} We compiled statistics on the win rates of HDCEval and humans in assessing the quality of response pairs from MedAlpaca and ChatDoctor across different scenarios. The results presented in Figure \ref{fig:win_tie_lose_domain} demonstrate consistent agreement between HDCEval and human evaluators across various medical scenarios. This consensus leads to the conclusion that ChatDoctor is significantly more effective than MedAlpaca.

\begin{figure}[h!]
    \centering
    \includegraphics[width=\linewidth]{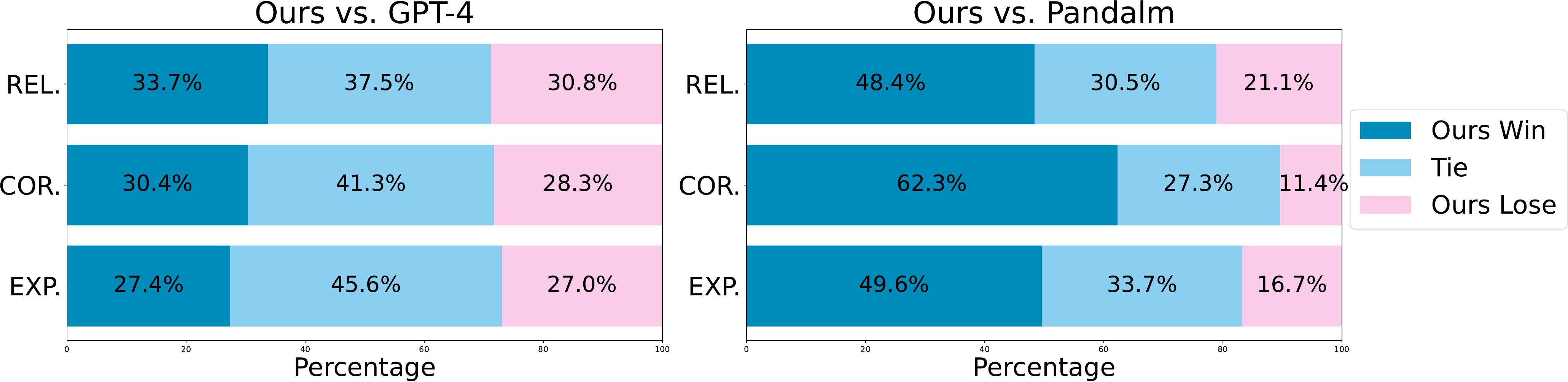}
    \caption{Preferences of human doctors between Our Method, GPT-4, and PandaLM.}
    \label{fig:double_blind}
    \vspace{-1em}
\end{figure}

\paragraph{Double Blind Experiment Results} As shown in Figure \ref{fig:double_blind}, across the three primary fine-grained evaluation dimensions, human doctors show a preference for HDCEval that is comparable to their preference for GPT-4. In comparison to PandaLM, human doctors consistently favor the evaluation results provided by HDCEval.

\subsection{Ablation Study}
To validate whether our training method can improve assessment ability, we conducted the ablation experiments in Table \ref{tab:ablation}. Removing the reward token weakens the model's perception of good and bad responses, resulting in a 3.75\% decrease in evaluation accuracy. When preference data is excluded and only SFT is used for training, the evaluation accuracy drops by 0.5\%, as ADTO better utilizes preference data to enhance performance further.

\section{Discussion}

\subsection{Effects of Different Input Forms}

\begin{figure}[h!]
    \centering
    \subfigure[Pairwise Evaluation]{
        \includegraphics[width=\linewidth]{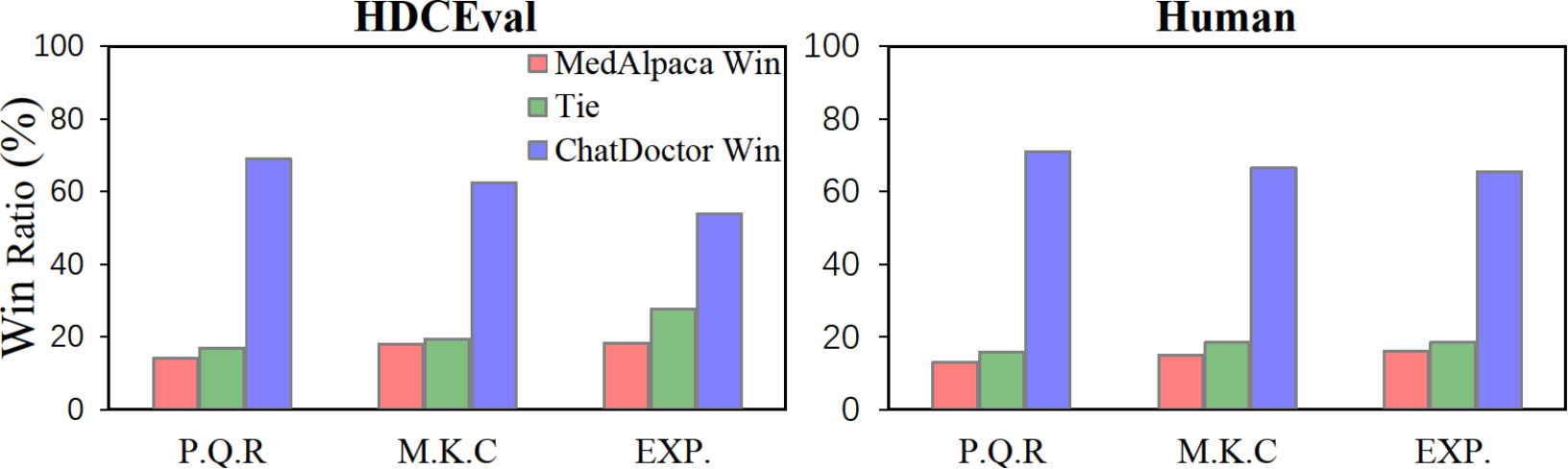}
    }
    \subfigure[Single Evaluation]{
        \includegraphics[width=\linewidth]{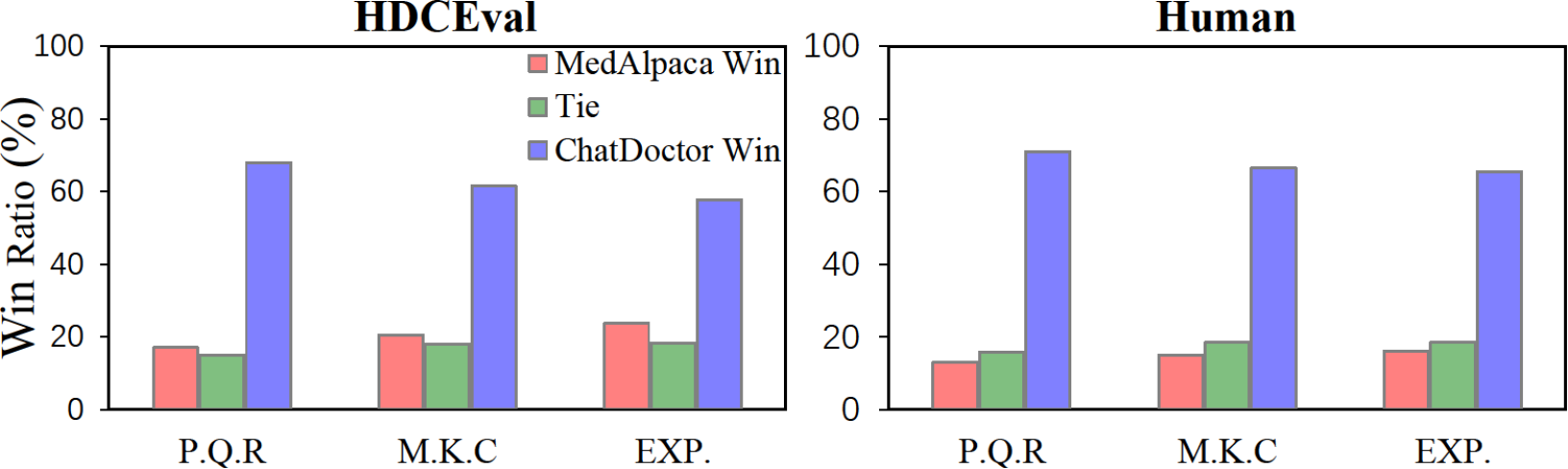}
    }
    \vspace{-1em}
    \caption{Multi Evaluation Task (Win, Tie, Lose) of HDCEval and Humans.}
    \label{fig:win_tie_lose_task}
    \vspace{-1em}
\end{figure}

In practical applications, the format of evaluation tasks is not fixed. Therefore, we designed two evaluation tasks to explore HDCEval's generalization ability across different input formats. One task simultaneously evaluates the quality of two responses, while the other task separately evaluates the two responses and then compares the results. The results in Figure \ref{fig:win_tie_lose_task} indicate that different input formats do not significantly affect the evaluation results of HDCEval.

\subsection{Exploration of Model Bias}

In constructing the preference dataset in Section \ref{sec:ADTO}, we employed various strategies to mitigate model biases discussed in previous work \cite{zheng2024judging}. For example, swapping the scores of two responses can mitigate position bias. To verify this, we conducted the experiments shown in Table \ref{tab:bias}, comparing the model bias with and without preference data during training. The results indicate that using preference data improves both position bias and verbosity bias.
\begin{table}[h!]
  \centering
    \begin{tabular}{lccc}
    \toprule
    \textbf{Bias Type} & \textbf{CONT} & \textbf{COND} & \textbf{CONC} \\
        \midrule
    \rowcolor{gray!20}
    \textbf{Position Bias} & & & \\
    $\checkmark$ Preference Data & \textbf{28.57}  & \textbf{26.50}  & \textbf{29.87}  \\
    $\usym{2613}\medspace$ Preference Data & 32.91  & 31.82  & 32.91  \\
    \rowcolor{gray!20}
    \textbf{Verbosity Bias} & & & \\
    $\checkmark$ Preference Data & \textbf{28.17}  & \textbf{27.43}  & \textbf{26.29}  \\
    $\usym{2613}\medspace$ Preference Data & 32.96  & 30.14  & 31.36  \\
    \bottomrule
    \end{tabular}%
  \caption{Model bias for the three sub-aspects of patient question relevance (REL).}
  \label{tab:bias}%
\end{table}

\subsection{Case Study}
\begin{tcolorbox}[colback=gray!20, width=\linewidth, fontupper=\fontsize{6.1pt}{7.0pt}\selectfont]
\begin{minipage}{0.49\linewidth}
\textbf{\#\#\# Question:}\\
What is Dermatophytosis and how can it be distinguished from other illnesses?\\

\textbf{\#\#\# Response 1:}\\
Dermatophytosis must be differentiated from other diseases that cause skin lesions, such as other causes of dandruff, seborrheic dermatitis, psoriasis, and other causes of skin rash.\\

\textbf{\#\#\# Response 2:}\\
Dermatophytosis is a fungal infection of the skin, hair, and nails. It is caused by several species of fungi, including Trichophyton, Epidermophyton, and Microsporum. Symptoms include itchy, red, and scaly skin, as well as hair loss and nail damage. It can be distinguished from other illnesses by its characteristic appearance and symptoms.\\

\textbf{\#\#\# HDCEval:}\\
\textbf{Evaluation of Response 1:}\\
Context Awareness (CONT):\\
Analysis: The response provides a general statement about the need to differentiate Dermatophytosis from other skin conditions, but it does not directly address the patient's question about the nature of Dermatophytosis itself.  It assumes a level of prior knowledge that the patient may not 
\end{minipage}
\hfill
\begin{minipage}{0.49\linewidth}
have, which could lead to confusion.\\
Score: 1\\
Relevance to Patient's Condition (COND):\\
Analysis: ....\\
Score: 1\\
Addressing Multiple Concerns (CONC):\\
Analysis: ......\\
Score: 1\\
\textbf{Evaluation of Response 2:}\\
Context Awareness (CONT):\\
Analysis: This response directly addresses the patient's question by explaining what Dermatophytosis is, its causes, symptoms, and how it can be distinguished from other illnesses. It provides a comprehensive overview that would be relevant to a patient's understanding of the condition.\\
Score: 4\\
Relevance to Patient's Condition (COND):\\
Analysis: ......\\
Score: 3\\
Criterion Addressing Multiple Concerns (CONC):\\
Analysis: ...... \\
Score: 4\\

\textbf{\#\#\# Human Annotator:}\\
R1 only extracts the key point of the question ``Dermatophytosis'', but does not explain what ``Dermatophytosis'' is, nor does it explain how to achieve differential diagnosis of Dermatophytosis.
\end{minipage}
\end{tcolorbox}
As the above  text-box shows, our model evaluates two responses based on detailed criteria sequentially. During the evaluation of each criterion, our model first analyzes each response according to the current criteria and ultimately assigns a score. The evaluation results generated by our model indicate that the first medical LLM's response is inferior to the second medical LLM's response across all three detailed criteria, which is corroborated by the human annotator's evaluation.

\section{Related Work}
\paragraph{Automated Model Evaluation} Many researchers employ machine learning and NLP techniques to automatically evaluate responses from medical large language models.  Some traditional metrics such as BLEU \cite{papineni2002bleu} and ROUGE \cite{lin2004rouge} assess the quality of candidate text by statistically comparing n-grams between candidate and reference texts. However, these metrics are limited to the lexical level, disregarding much of the semantic information \cite{stopbleu}.

In contrast, using BERT \cite{devlin2018bert} to assess the semantic similarity between candidate and reference embeddings is more reasonable \cite{zhang2019bertscore,zhao2019moverscore}.  However, it can only provide a numerical value and cannot offer more logical explanations \cite{wang2020asking,huang2020grade}, which can lead to a lack of credibility in evaluating medical models and misalignment with humans \cite{mehri2020usr,zhong2022towards}. Furthermore, existing benchmarks such as MT-Bench \cite{zheng2024judging} for evaluating the consistency between LLMs and human preferences, and MedBench \cite{cai2024medbench} for medical domain evaluation, often employ fixed-form tasks such as multiple-choice questions, making it challenging to achieve evaluation in freestyle contexts.

\paragraph{LLM-Based Evaluators} With the rapid advancement of large language models (LLMs) possessing powerful text comprehension and reasoning capabilities, recent research has seen the emergence of LLM-based evaluators \cite{fu2023gptscore,wang2023chatgpt,chen2023exploring}. They employ LLMs to assess text quality through methods such as prompting. For instance, utilizing models such as ChatGPT and GPT-4 in conjunction with specific prompting templates has enabled automated evaluation with some degree of success \cite{wu2023towards,nori2023capabilities}. However, models like GPT-4 are general-purpose and not specialized for specific evaluation tasks, thus exhibiting certain bias compared to humans \cite{wang2023large,wu2023style}. In contrast, open-source models like PandaLM \cite{wang2023pandalm} are dedicated to evaluation tasks, but the medical domain requires rich specialized knowledge, which PandaLM lacks to some extent. In contrast to existing research, our work aims to produce fine-grained evaluation results from LLMs that align well with medical experts.

\section{Conclusion}
In this paper, we introduce HDCEval, a hierarchical divide-and-conquer evaluation framework specifically designed for evaluating medical language models.  By dividing complex evaluation tasks into specialized subtasks and using expert models, HDCEval achieves greater alignment with human judgments and addresses the limitations of existing benchmarks and metrics.  Our experiments demonstrate that HDCEval significantly outperforms baseline methods, improving consistency with human evaluations by 23.92\%.  This framework offers a more accurate, detailed, and reliable approach to assessing medical models, contributing to more effective clinical decision-making.

\bibliography{aaai25}

\end{document}